\documentclass{article}

\usepackage{pifont}
\usepackage[table,xcdraw]{xcolor}
\usepackage{tcolorbox}
\usepackage{amssymb}
\usepackage{tikz}

\usepackage[final]{neurips_2024}

\usepackage[utf8]{inputenc} 
\usepackage[T1]{fontenc}    
\usepackage{hyperref}       
\usepackage{url}            
\usepackage{booktabs}       
\usepackage{amsfonts}       
\usepackage{nicefrac}       
\usepackage{microtype}      

\usepackage{amsmath}        
\usepackage{graphicx}       
\usepackage{svg}            
\usepackage[normalem]{ulem} 
\usepackage{makecell}
\usepackage{setspace}
\usepackage{multirow}
\usepackage{subfigure}
\usepackage{amssymb}
\usepackage{times}
\usepackage{latexsym}
\usepackage[T1]{fontenc}
\usepackage[utf8]{inputenc}
\usepackage{microtype}
\usepackage{inconsolata}
\usepackage{amsmath}
\usepackage{amssymb}
\usepackage{graphicx}
\usepackage{enumerate}
\usepackage{balance}
\usepackage{blindtext}
\usepackage{listings}
\usepackage{booktabs}
\usepackage{colortbl}
\usepackage{tcolorbox}
\usepackage{marvosym}
\usepackage{float}
\usepackage{makecell}
\usepackage{multicol}
\usepackage{multirow}
\usepackage{tabularx}
\usepackage{marvosym}
\usepackage{stfloats}
\usepackage{graphicx}
\usepackage{subfigure}
\usepackage{subcaption}
\usepackage{ulem}
\usepackage{caption}
\usepackage{array}
\usepackage{makecell}
\usepackage{ragged2e}
\usepackage{hyperref}  
\usepackage{textcomp}    
\usepackage[utf8]{inputenc}
\usepackage{booktabs}
\usepackage{multirow}
\usepackage{tabularx}

\definecolor{lightgray}{gray}{0.9}

\definecolor{myblue}{RGB}{25, 120, 180}

\renewenvironment{quotation}
  {\list{}{\leftmargin=0em \rightmargin=0em}\item\relax}
  {\endlist}

\newcolumntype{C}[1]{>{\centering\arraybackslash}m{#1}} 
\newcolumntype{L}[1]{>{\raggedright\arraybackslash}m{#1}} 

\newcommand\footnoteONLYtext[1]{
    \let \mybackup \thefootnote
    \let \thefootnote \relax
    \footnotetext{#1}
    \let \thefootnote \mybackup
    \let \mybackup \imareallyundefinedcommand}

\lstdefinestyle{PythonCode}{
    language=Python,
    basicstyle=\ttfamily,
    breaklines=true,
    keywordstyle=\bfseries\color{NavyBlue},
    morekeywords={},
    emph={self},
    emphstyle=\bfseries\color{Rhodamine},
    commentstyle=\itshape\color{black!50!white},
    stringstyle=\bfseries\color{PineGreen!90!black},
    columns=flexible,
}

\lstdefinestyle{BashCode}{
    language=Bash,
    basicstyle=\ttfamily\color{white}, 
    backgroundcolor=\color{black},      
    breaklines=true,
    keywordstyle=\bfseries\color{MidnightBlue},
    morekeywords={},
    emph={},
    emphstyle=\bfseries\color{Purple},
    commentstyle=\itshape\color{black!50!white},
    stringstyle=\bfseries\color{OliveGreen!90!black},
    columns=flexible,
}

\lstdefinelanguage{json}{
    basicstyle=\ttfamily\small,
    showstringspaces=false,
    morestring=[b]",
    stringstyle=\color{blue},
    commentstyle=\color{darkgray}\ttfamily\itshape,
    numbers=left,
    numberstyle=\tiny\color{gray},
    stepnumber=1,
    numbersep=10pt,
    tabsize=2,
    breaklines=true,
    backgroundcolor=\color{lightgray},
    literate=
     *{0}{{{\color{purple}0}}}{1}
      {1}{{{\color{purple}1}}}{1}
      {2}{{{\color{purple}2}}}{1}
      {3}{{{\color{purple}3}}}{1}
      {4}{{{\color{purple}4}}}{1}
      {5}{{{\color{purple}5}}}{1}
      {6}{{{\color{purple}6}}}{1}
      {7}{{{\color{purple}7}}}{1}
      {8}{{{\color{purple}8}}}{1}
      {9}{{{\color{purple}9}}}{1}
      {:}{{{\color{black}:}}}{1}
      {,}{{{\color{black},}}}{1}
      {\{}{{{\color{black}\{}}}{1}
      {\}}{{{\color{black}\}}}}{1}
      {[}{{{\color{black}[}}}{1}
      {]}{{{\color{black}]}}}{1},
}

\lstdefinestyle{JsonCode}{
    language=json,
    basicstyle=\ttfamily\small,
    keywordstyle=\color{purple}\bfseries,
    stringstyle=\color{blue},
    commentstyle=\color{darkgray}\ttfamily\itshape,
    numbers=left,
    numberstyle=\tiny\color{gray},
    stepnumber=1,
    numbersep=10pt,
    tabsize=2,
    breaklines=true,
    backgroundcolor=\color{white},
    showspaces=false,
    showtabs=false,
    xleftmargin=10pt,
    frame=none,
    framesep=2pt,
    rulecolor=\color{black}
}

\title{Autonomous Agents for Collaborative Task under Information Asymmetry}

\author{
    \textbf{Wei Liu}{\footnotesize $^\bigstar$\textsuperscript{\textdagger}} \quad
    \textbf{Chenxi Wang}{\footnotesize $^\bigstar$\textsuperscript{\textdagger}} \quad
    \textbf{Yifei Wang}{\footnotesize $^\bigstar$} \quad
    \textbf{Zihao Xie}{\footnotesize $^\bigstar$} \quad
    \textbf{Rennai Qiu}{\footnotesize $^\bigstar$} \quad
    \textbf{Yufan Dang}{\footnotesize $^\bigstar$} \quad \\
    \textbf{Zhuoyun Du}{\footnotesize $^\bigstar$} \quad
    \textbf{Weize Chen}{\footnotesize $^\bigstar$} \quad
    \textbf{Cheng Yang}{\footnotesize $^\clubsuit$\textsuperscript{\Letter}} \quad
    \textbf{Chen Qian}{\footnotesize $^\bigstar$\textsuperscript{\Letter}} \quad \\
  {\footnotesize $^\bigstar$}Tsinghua University \quad
  {\footnotesize $^\clubsuit$}Peng Cheng Laboratory, China \quad 
  \\
  \texttt{thinkwee2767@gmail.com} \quad 
  \texttt{albertyang33@gmail.com} \quad 
  \texttt{qianc62@gmail.com} \quad 
}

\begin{document}
\maketitle
\vspace{-3em}
\begin{center}
    {\large \textcolor{myblue}
    {\href{https://thinkwee.top/iagents/}{\texttt{https://thinkwee.top/iagents/}}}}    
\end{center}

\begin{abstract}
Large Language Model Multi-Agent Systems (LLM-MAS) have greatly progressed in solving complex tasks.
It communicates among agents within the system to collaboratively solve tasks, under the premise of shared information.
However, when agents' collaborations are leveraged to perform multi-person tasks, a new challenge arises due to information asymmetry, since each agent can only access the information of its human user. Previous MAS struggle to complete tasks under this condition.
To address this, we propose a new MAS paradigm termed \textit{iAgents}, which denotes \textit{\textbf{I}nformative Multi-\textbf{Agent} \textbf{S}ystems}. In \textit{iAgents}, the human social network is mirrored in the agent network, where agents proactively exchange human information necessary for task resolution, thereby overcoming information asymmetry. \textit{iAgents} employs a novel agent reasoning mechanism, \textit{InfoNav}, to navigate agents' communication towards effective information exchange. 
Together with \textit{InfoNav}, \textit{iAgents} organizes human information in a mixed memory to provide agents with accurate and comprehensive information for exchange.
Additionally, we introduce \textit{InformativeBench}, the first benchmark tailored for evaluating LLM agents' task-solving ability under information asymmetry. Experimental results show that \textit{iAgents} can collaborate within a social network of 140 individuals and 588 relationships, autonomously communicate over 30 turns, and retrieve information from nearly 70,000 messages to complete tasks within 3 minutes\footnote{Available on \url{https://github.com/thinkwee/iAgents}.}.
\end{abstract}

\begin{quotation}
\begin{justify}
``\textit{A friend is someone with whom there is mutual understanding, emotional support, and shared experiences.}''
\end{justify}
\begin{flushright}
----- Joey's and Chandler's agents discuss the word "friend", \\after experiencing the whole \textit{Friends} season one story.
\end{flushright}
\end{quotation}

\footnoteONLYtext{\textdagger: Equal Contributions.}
\footnoteONLYtext{\Letter: Corresponding Authors.}

\vspace{-1em}

\section{Introduction}

\setlength{\textfloatsep}{5pt plus 1.0pt minus 2.0pt}
\begin{figure*}[ht]
    \centering
    \includegraphics[width=0.99\linewidth]{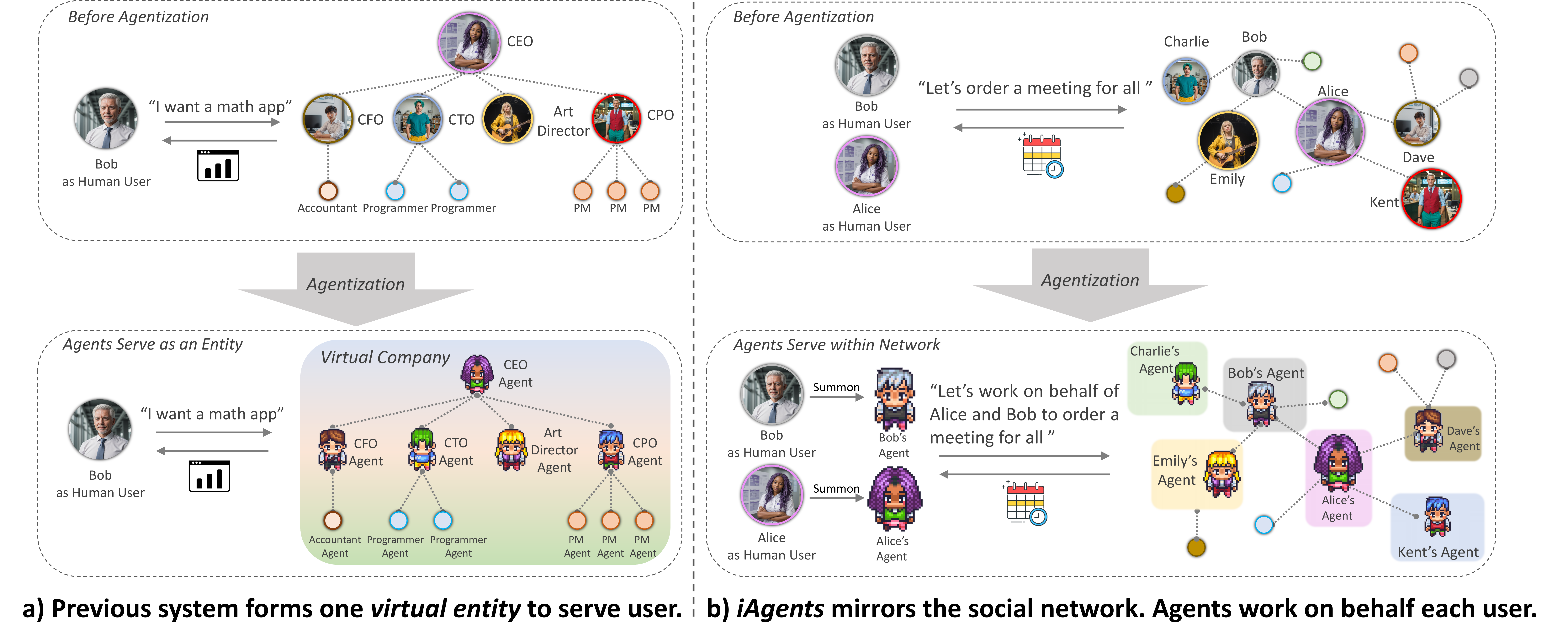}
    \caption{Comparison between previous MAS (left) and \textit{iAgents} (right). The visibility range of information for each agent is highlighted with a colored background. On the left, all agents share all information (colored background of \textit{Virtual Company}). On the right, each agent could only see information about its human user (separated colored backgrounds), and \textit{iAgents} is designed to deal with such kind of information asymmetry.} 
    \label{fig:scheme_difference}
\end{figure*}

There has been notable progress in autonomous agents driven by the Large Language Model (LLM), especially in developing communicative agents for completing collaborative tasks~\cite{xi2023rise,wang2024survey,guo2024large,qian2023communicative, hong2023metagpt, tang2023medagents, li2024agent, zheng2023chatgpt, qian-etal-2024-experiential}, as shown in Figure~\ref{fig:scheme_difference}a.
In these multi-agent systems (MAS), multiple agents are created through role-play prompting~\cite{li2024camel} to imitate the ability of human experts and form a \textit{virtual entity} (e.g., an agent company or hospital) to provide solutions derived from agents' communication. Agents share context in the \textit{virtual entity} to facilitate collective decision-making.

Since autonomous communication among agents has achieved significant success in discussing, decomposing, and resolving various complex tasks, the natural idea is to upgrade the tasks for agents from single-person to multi-person, where agents work on behalf of multiple human users and solve the collaboration task among these users.
An intuitive solution is to assign each user an agent and perform autonomous collaborations among these agents. However, in such a setting, a new challenge arises. This challenge involves dealing with asymmetry~\cite{Weber_1978, tomasello2009cultural} in various types of information (environment, goals, and mind state)~\cite{zhou2024sotopia, zhou2024real, premack1978does, bubeck2023sparks, xu2024opentom, chen2024internet} since each agent can only observe the information of its human user. Previous LLM-MAS are not suitable for handling this scenario, because
1) human information is sensitive and private, so the asymmetry can not be resolved by directly collecting all information into one place and sharing it as the context for MAS.
2) Human information is dynamic so it can not be easily memorized during pre-training and activated accurately through role-play prompting in MAS to avoid asymmetry.
Essentially, agents' cooperation in previous MAS has adopted an introspective approach within the \textit{virtual entity} (an agent hospital/town/software company), which struggles to deal with asymmetry in human information.

To bridge gaps in such asymmetry, agents need to retrieve information from humans and proactively exchange information, creating a new ecosystem combining the human and the agent network.
Therefore, we propose the concept of \textit{iAgents} (\textit{\textbf{I}nformative Multi-\textbf{Agent} \textbf{S}ystems}) for achieving this kind of collaboration, as shown in Figure~\ref{fig:scheme_difference}b. \textit{iAgents} utilizes a new agent reasoning method (\textit{InfoNav}) to model the agents' minds and navigate communication among agents toward proactive information exchange. Furthermore, a new memory mechanism is designed to provide agents with accurate and comprehensive information for exchange. Additionally, we introduced \textit{InformativeBench}, the first benchmark evaluating agents' collaboration ability under information asymmetry. It includes both information-seeking tasks within large social networks and algorithm-like reasoning tasks over a small network. Our contributions can be summarized as follows:
\begin{enumerate}
    \item We raise the research problem of information asymmetry in the multi-agent system for enhancing human collaboration, which is the first to shift the research perspective in this area from a holistic system view to individuals within the system. It gives a new vision to the human-agent collaboration relationship.
    \item We propose the \textit{iAgents} framework to deal with the information asymmetry in a multi-agent system. Equipped with \textit{InfoNav} and improved memory mechanism, \textit{iAgents} could perform effective communication and collaboration within a social network (shown in Figure~\ref{fig:network_dis}) of 140 individuals and 588 relationships, and across over 30 dialogues they searched nearly 70,000 messages and resolved the task within 3 minutes.
    \item We introduce the first multi-agent information asymmetry benchmark, \textit{InformativeBench}. Agents with some state-of-the-art LLM backends achieved an average accuracy of 50.48\% on \textit{InformativeBench}, with the most challenging task achieving only 22.8\% accuracy, which reveals both potential promise and challenges in this direction.
\end{enumerate}

\section{Related Work}
\textbf{Agents based on Large Language Models (LLMs)} Originate from ancient Greek philosophy, where an "agent" denoted a being capable of intentional action, driven by mental states such as desires and beliefs~\cite{sep-agency}. As AI progresses, this concept integrates into the simulation and understanding of intelligent behavior~\cite{Wooldridge_Jennings_1995}. Traditionally, agent research focused on some specific tasks~\cite{10.1109/64.180407, 10.1145/176789.176804, sutton1999reinforcement}, but the emergence of LLMs has shifted this focus. GPT-4, for instance, is recognized for achieving a form of general intelligence~\cite{bubeck2023sparks}, prompting exploration into equipping LLMs with agency and intrinsic motivation~\cite{xi2023rise}. Studies introduce frameworks for LLM-based agents, including memorisation~\cite{zhang2024survey}, decision-making, perception, and action modules, leveraging the inherent autonomy, reactivity, pro-activeness, and social ability of LLMs~\cite{weng2023agent}. Notable applications include utilizing single-agent systems and multi-agent systems for task solving and simulation~\cite{qian2023communicative, hong2023metagpt, tang2023medagents, li2024agent, zheng2023chatgpt, chen2024agentverse, park2023generative, qian2024iterative}. However, most of this research has primarily focused on agent capabilities, often neglecting interaction and cooperation paradigms. This highlights a critical research gap~\cite{boiko2023emergent}, warranting further exploration in this area.

\textbf{Human-Agent and Multi-Agent Cooperation Paradigms} To ensure that agents align with human objectives~\cite{kenton2021alignment, paul2023}, human-agent cooperation is crucial. Two main paradigms of human-agent cooperation are the Equal-Partnership and the Instructor-Executor. The former emphasizes agents as communicators who understand human emotions~\cite{Hasan_2023}, while the latter highlights the human's guiding role, with agents following instructions~\cite{gao2023assistgpt}. However, single-agent systems face limitations, such as the inability to collaborate, learn from social interactions, and function effectively in complex scenarios~\cite{Significant_Gravitas_AutoGPT,liu2023training,xi2023rise}.
Research suggests that multi-agent systems with each agent holding specific functions, can stimulate stronger intelligence~\cite{minsky1988society, Balaji2010AnIT, qian2024scaling}. Collaborative multi-agent systems can efficiently handle complex tasks~\cite{mandi2023roco, liang2023encouraging, du2024multi}. 
Recent studies focus on scenarios with information asymmetry among agents. For instance, ~\cite{zhou2024sotopia} explores social intelligence among agents achieving private goals based on common scene information. Additionally, ~\cite{zhou2024real} develops an evaluation framework to simulate social interactions with LLMs. The study finds that learning from omniscient simulations enhances interaction naturalness but doesn't improve goal achievement in cooperative scenarios. Many real-world scenarios involve information asymmetry, posing challenges for multi-agent systems.

\textbf{Reasoning} In previous research, agents are tasked with providing accurate information to human users in human-machine collaboration. Due to the nature of language models, the output of information relies on the user's input and the previously decoded content serving as rationale context. Therefore, some work on reasoning has explored how to improve the organization and expression of this rationale context~\cite{NEURIPS2022_9d560961, NEURIPS2023_271db992, besta2024graph, ding2023everything}, to enhance the accuracy of output information. However, some research has also found that the reasoning process of LLMs differs from that of humans~\cite{yang2024large, jin2024impact, chen2024masked, pfau2024let}. For questions relying on internal knowledge within LLMs to answer, agents do not necessarily solve problems step by step like humans. Instead, compared to steps, having context with sufficient information content is more important. For scenarios discussed in this paper, we focus on machine-to-machine communication, relying on external knowledge of LLMs to collaboratively answer questions. This poses different requirements for LLM reasoning, especially in terms of how to promote information flow so sufficient information is included in the context provided to LLM. Agents need to actively~\cite{hu2024uncertainty} and accurately acquire, provide, and ask for information.
    
\section{Method} \label{method}
\begin{figure*}[t]
    \centering
    \includegraphics[width=0.99\linewidth]{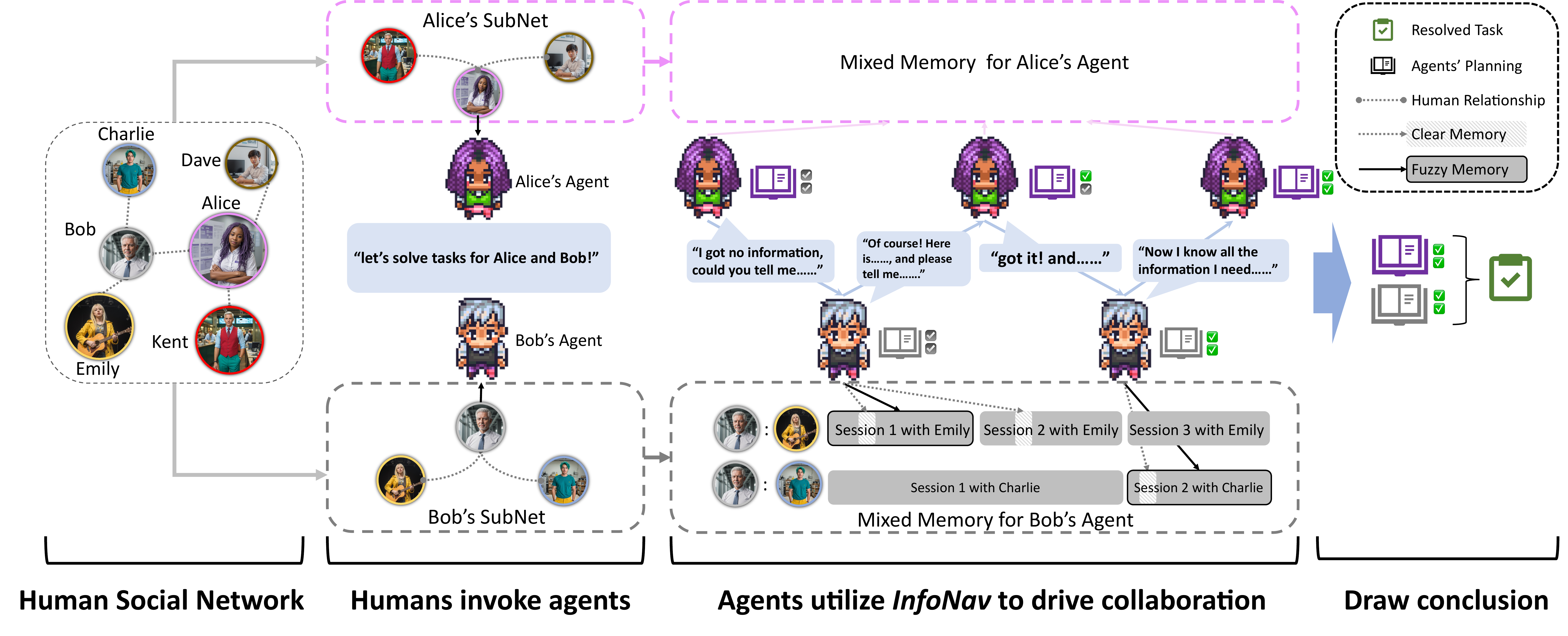}
    \caption{Overall architecture of \textit{iAgents}. From left to right, 1) each individual in the social network is equipped with an agent, and 2) two human users invoke their agents to solve a task, each initially holding the information that is visible to its human user. Then 3) agents automatically raise communication and exchange necessary information on behalf of human users. Finally, 4) agents perform a consensus check on their planning completed by \textit{InfoNav} to solve the task.}
    \label{fig:overall_arch}
\end{figure*}

\subsection{Problem Formulation}
Without loss of generality, we formalize tasks in social networks that require information exchange for collaboration as a Question Answer (QA) task. The rationales $R$ necessary for answering the question $Q$ are distributed in different human information ($I_1, I_2$) across the social network, which leads to information asymmetry. Consequently, agents ($A_1, A_2$) of two individuals are required to collaborate, update the rationale set ($R_1, R_2$) that they hold through communication $C$, and by combining their rationales, they can reason and obtain the answer. The whole process can be formulated as:
\begin{align}
       Ans &= Reasoning(Q, R)  \label{eq:prob_define_1} \\
       R &= R_1 \cup R_2 \label{eq:prob_define_2} \\
       R_1, R_2 &= C(I_1, I_2, A_1, A_2)  \label{eq:prob_define_3} 
\end{align}

\subsection{Overview}

As shown in Figure~\ref{fig:overall_arch}, agents need to actively retrieve information from humans and exchange it with other agents. The communication can be represented as:
\begin{equation}
C_n = \{ U_1, U_2, ..., U_n \}
\end{equation}
where $U$ denotes an utterance in the communication $C$, and $n$ is the maximum number of communication turns. Agents take turns making utterances to advance towards task resolution.
Following the classical definition~\cite{yao2023react, xi2023rise}, where agents observe the environment, think to make decisions, and then take action, we can organize agents' communication similarly. Each agent's behavior in one communication turn involves a pipeline of 1) \textit{observing} the current communication progress $C$ and their held rationales $R$, 2) \textit{thinking} about how to update the rationale to $R^{new}$ and what $query$ to make for retrieving information from humans, and 3) \textit{acting} by retrieving information and making an utterance based on it. This pipeline can be formalized as:

{\small
\begin{equation}
U_i =
\left\{
\begin{array}{ll}
    \text{$Act_{A_1}(Think_{A_1}(Obs_{A_1}^{i}))$} & \text{$i \% 2 == 1$} \\
    & \\
    \text{$Act_{A_2}(Think_{A_2}(Obs_{A_2}^{i}))$} & \text{$else$}  \label{eq:react}
\end{array}
\right.
\end{equation}
}
where
\begin{align}
    Obs_{A}^{i} &= \{ R, C_{i-1} \} \label{eq:observation} \\
    Think_{A}(Obs_{A}^{i}) &= \{ query, R^{new} \} \label{eq:think} \\
    Act_{A}(Think_{A}) &= A(query(I)) = U \label{eq:act}
\end{align}
To ensure each generated utterance provides valuable information and eliminates asymmetry, how to exchange information and what information to exchange is crucial. To deal with these two questions, we use the \textit{\textbf{InfoNav}} mechanism to guide communication towards effective information exchange. Furthermore, we introduce the \textit{\textbf{Mixed Memory}} mechanism which organizes human information into Fuzzy and Clear Memory for accurate and comprehensive retrieval. Additionally, each agent can initiate new communication $C^{new}$ within their subnetwork, which means the communication $C$ may be recursive and can diffuse among the social network:
\begin{align}
    C_n &= \{ C^{new}_1, C^{new}_2, ..., C^{new}_m, U_1, U_2, ..., U_n \} 
\end{align}
For example, if Alice's agent wants to collaborate with Bob's agent, Bob's agent might respond "Hold on, I can ask Charlie's agent for help."

\subsection{InfoNav}
\begin{figure*}[t]
    \centering
    \includegraphics[width=0.99\linewidth]{Fig/infonav_case.pdf}
    \caption{A case of the task asking two agents to find the longest activity among all schedules. \textit{InfoNav} navigates the communication by providing a plan to the agent. It first 1) asks the agent to make a plan on what information is needed, then 2) fills the placeholder in this plan during communication. Finally it 3) performs a consensus check on the completed plan to 4) get the answer.}
    \label{fig:InfoNav_case}
\end{figure*}
As shown in Equation~\ref{eq:observation}, the agent needs to be aware of its rationale set and ongoing communication to effectively advance the conversation. While the status of the rationale set can be implicitly inferred from utterances, this inference is often unreliable for LLM Agents. This unreliability can lead to incorrect states and then generate meaningless utterances, such as repetitive questioning or redundant thanking, which makes it harder to infer rationale and creates a vicious cycle. 
To address this, we propose the \textit{InfoNav} mechanism. \textit{InfoNav} plans and tracks the status of the agent's rationale set explicitly for better navigating the communication.
Before each utterance, the agent reviews its plan to identify which unknown rationale to inquire about and then updates the plan based on the responses received. Figure~\ref{fig:InfoNav_case} shows an example of \textit{InfoNav} in action. Initially, we prompt the agent to generate a plan $P$ outlining the rationales needed to answer question $Q$. Since the agent has no information at the beginning, all rationales in the plan are marked as unknown:
\begin{align}
    P(r_1^{u}, ..., r_m^{u}) &= Prompt(Q)
\end{align}
where $r^u$ denotes unknown rationales. During communication, if the agent gets the information of one rationale, it will update the status of this rationale from ``unknown'' to ``known'' and fill this information into the rationale placeholder in the planning text. The plan is written in fluent natural language, making it explicit and effective for prompting the model. Therefore, using \textit{InfoNav}, the rationale set $R$ in equation~\ref{eq:observation} is rewritten to plan $P$, and the updated rationale $R^{new}$ is replaced to the plan with filled rationales $P(r^{k})$, where $r^k$ represents known rationales:
\begin{align}
    Obs_{A} &= \{ P(r^{u}), C \} \\
    P(r^{k}) &= Think_{A}(Obs_{A}) 
\end{align}
After multiple turns of communication, both sides finish the update of their plans. Agents then unify collected rationales and discard conflicting ones to reach an answer, denoted as ``Consensus Reasoning''. Thus, equations~\ref{eq:prob_define_1} to \ref{eq:prob_define_3} rewrite to:
\begin{align}
    Ans &= Reasoning(Q, R) \\
    R &= Consensus(P_1(R_1), P_2(R_2)) \\
    P_1(R_1), P_2(R_2) &= C(I_1, I_2, A_1, A_2)
\end{align}
Previous reasoning methods\cite{NEURIPS2022_9d560961, NEURIPS2023_271db992, besta2024graph, ding2023everything} focused on providing accurate plans. In contrast, \textit{InfoNav} emphasizes navigating communication and information exchange with plans. The plan in \textit{InfoNav} can be seen as a generalization of Dialogue Status Tracking (DST)\cite{henderson2014word, wu2019transferable, heck2020trippy} in conventional task-oriented dialogue systems. It also generalizes the concept of software in multi-agent software generation frameworks like ChatDev\cite{qian2023communicative} or MetaGPT~\cite{hong2023metagpt}. The plan maintains progress in task-solving, guiding agents to share information during communication.

\subsection{Mixed Memory}
In \textit{iAgents}, agents are navigated by \textit{InfoNav} to retrieve human information and share it with other agents for collaboration. Retrieval of human information is necessary since 1) human's lifelong information can not be stored in the ``long context'' (such as 128k tokens) of LLM, and 2) even though the information required for a single-turn conversation can fit into the context, the accumulation of information over multiple turns can lead to context explosion. It is also challenging to organize human information which is diverse in format and complex to understand. We propose organizing human information into two types of agent memories: Clear Memory and Fuzzy Memory. These memories facilitate reactive retrieval, ensuring accurate and comprehensive rationale extraction, as shown in Figure~\ref{fig:overall_arch}.

Clear Memory ($Mem_C$) stores information in a structured format to facilitate precise retrieval. Clear memory faithfully preserves the original information ($I$) and supports accurate retrieval. Additionally, it enables information retrieval from multiple spans across different chat sessions ($s$), capturing evolving changes in rationales.

However, Clear Memory's strict exact-match requirements complicate the retrieval process. It also struggles to provide cohesive context. To address these issues, we introduce Fuzzy Memory ($Mem_F$). Fuzzy memory stores summarized session texts ($I_s$) and uses embedding-based ANN retrieval~\cite{johnson2019billion}. Although both fuzzy memory and reflection~\cite{park2023generative} produce summary-like text, we emphasize objective summarization of information to facilitate session-level retrieval, rather than subjective generalizations to aid in planning. While this approach may lose some details, it offers a comprehensive context and enables robust, semantic-based retrieval. Therefore, the retrieval action in Equation~\ref{eq:act} can be rewritten to involve both memory types:
\begin{align}
    Act_{A_k}(Think_{A_k}) &= A_k(query_k(I_k)) \\
    &= A_k(SQL(Mem_C), ANN(Mem_F)) 
\end{align}
What's more, the query of these two kinds of memories is decided by agents based on observations of previous executions, which means agents can reactively adjust their queries. 
Combining these two kinds of memory facilitates agents to cross-verify the retrieved information and provides \textit{InfoNav} with comprehensive and accurate rationales.

\section{InformativeBench} \label{benchmark}

\begin{figure*}[ht]
    \centering
    \includegraphics[width=0.9\linewidth]{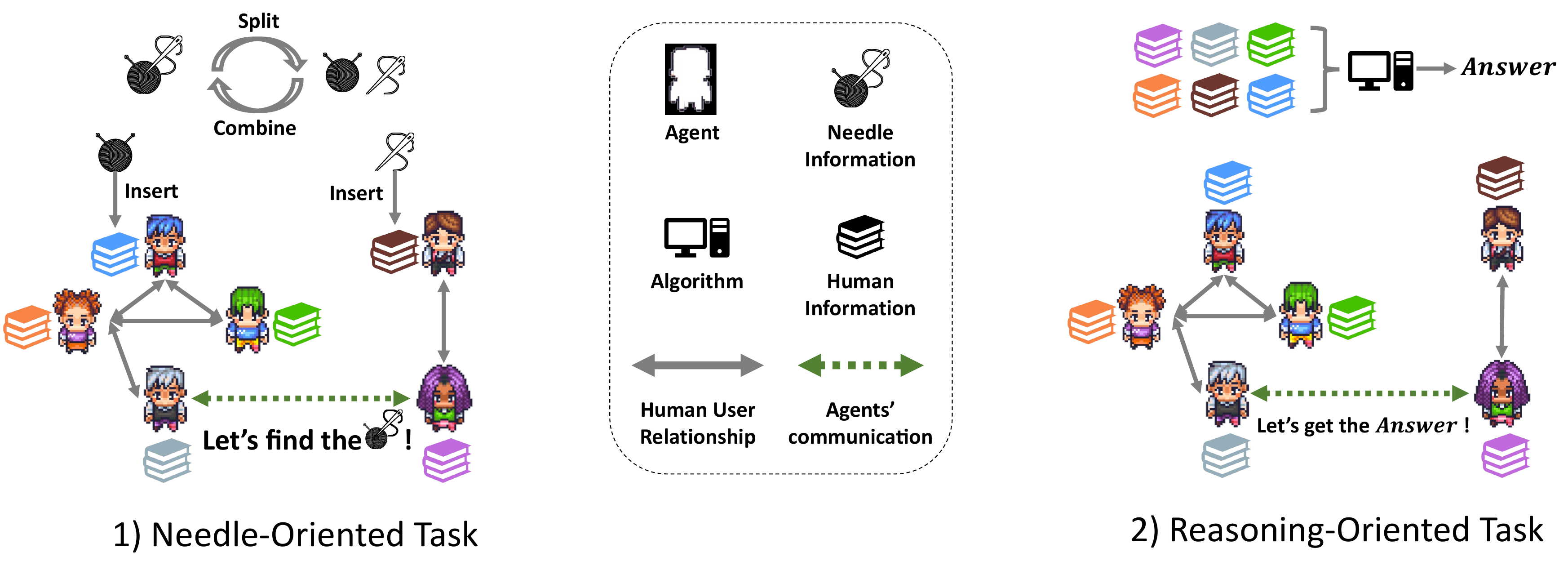}
    \caption{Two kinds of tasks in the \textit{InformativeBench}. Each agent can only see the information (marked with different colors) of the human that it works on behalf of, which generates information asymmetry. Agents are 1) asked to find the needle information within the network or 2) reason to get an answer which is the output of an algorithm running on distributed information in the network.}
    \label{fig:task}
\end{figure*}

To the best of our knowledge, there is no benchmark or dataset tailored for information asymmetry in the collaboration task among communicative agents. 
In this paper, we construct \textit{InformativeBench}, the first benchmark to evaluate agent collaboration tasks featuring information asymmetry in social networks. It includes two categories with a total of five datasets. 
Details, including the scale, distribution, and metrics of the datasets, are provided in section~\ref{appendix:bench_details}.
What's more, recent studies have found that \textit{LLM continuously ingests internet data so static benchmarks can be easily memorized and overfitted}~\cite{zhou2023don, xu2024benchmarking, zhang2024careful}. 
Hence, two pipelines for constructing \textit{InformativeBench} are easy to realize and can be generalized to more domains for constant and dynamic evaluations. They are Needle-Oriented and Reasoning-Oriented pipelines, as shown in Figure~\ref{fig:task}.

\textbf{Needle-Oriented Pipeline} A "needle"\cite{kamradt2023needle} is inserted into the social network, and agents are tasked with finding this "needle" information. This evaluates their ability to share and locate information. The dataset can be created by splitting the needle and spreading it into the network, or by collecting pieces from the network and combining them. For the split method, the \textbf{\textit{Needle in the Persona (NP)}} dataset modifies the dialogue in the SPC dataset\cite{jandaghi2023faithful} by adding a common or opposite persona to two individuals' personas. Agents are asked to find this persona. For the combination method, the \textbf{\textit{FriendsTV}} dataset reconstructs the social network from the entire Season 1 script of \textit{Friends}~\cite{wiki:Friends_TV_series}, involving 140 characters with 588 relationships, and combines two questions in the FriendsQA dataset~\cite{yang2019friendsqa, li2020transformers} as "needle pieces" to generate new question. This dataset, the largest in \textit{InformativeBench}, features sarcasm, plot twists, and complex relationships for simulating real-world challenges.

\textbf{Reasoning-Oriented Pipeline} Humans are assigned different pieces of information, which serve as inputs for an algorithm (such as sorting or merging). Agents must reason to get the answer which is the algorithm's output. Therefore, the algorithm serves as an automatic verifier for information asymmetric reasoning. In \textit{InformativeBench}, this is represented by the \textbf{\textit{Schedule}} dataset, which develops a program for assigning different schedules to individuals. Agents are presented with algorithmic problems of varying difficulties, and the program automatically verifies the correctness of their solutions. The datasets include questions of three levels of difficulty: \textit{Easy}) calculate the number of conflicting schedules between two people, \textit{Medium}) find the longest activity among six people, and \textit{Hard}) find the longest common free period among six people.

\section{Experimental Setup} \label{experimental_setup}
We generically treat chat histories as human information. This approach simplifies modeling information asymmetry in social networks. Other types of information, such as knowledge bases, documents, or web content, can all be organized in mixed memory so \textit{iAgents} is adaptable to all these kinds of information.
We conduct all experiments with a maximum of 10 communication turns for agents. The experiments use gpt-4-0125-preview, gpt-3.5-turbo-16k, gemini-1.0-pro-latest, and claude-sonnet~\footnote{as of 20240501.} as LLM backends. The temperature is set to 0.2. For Fuzzy Memory, we use gpt-4-0125-preview to summarize session text and OpenAI text-embedding-3-small to generate embeddings for ANN embedding search.
We use precision as the metric for questions in the NP, ScheduleEasy, and FriendsTV datasets. For the ScheduleMedium and ScheduleHard datasets, we use F1 and IoU as the metrics, corresponding to the algorithm used. Details about the metrics are shown in Section~\ref{appendix:metrics}. 
For the Schedule and NP datasets, we do not activate mixed memory since the information scale is small and can be fully loaded in the LLM context. Additionally, for the Schedule dataset, we do not activate the agent's ability to initiate new communication due to the small scale of the social network.

\section{Result}
\subsection{\textit{InformativeBench} Evaluations}

\begin{table}[h]
\centering
\small
\begin{tabular}{c|ccc|cc}
\toprule[1.5pt]
\multirow{2}{*}{\textbf{LLM Backend}} & \multicolumn{3}{c|}{\textbf{Reasoning-Oriented (Schedule Dataset)}} & \multicolumn{2}{c}{\textbf{Needle-Oriented}} \\
           & \textbf{Easy} & \textbf{Medium} & \textbf{Hard} & \textbf{NP}      & \textbf{FriendsTV} \\ \midrule[1pt]
GPT 4      & 56.67\%      & 51.00\%        & 22.80\%      & 64.00\% & 57.94\%   \\
GPT 3.5    & 36.67\%      & 18.00\%        & 12.25\%      & 51.00\% & 35.71\%   \\
Claude Sonnet     & 43.33\%      & 17.44\%        & 18.66\%      & 50.00\% & 34.13\%   \\ 
Gemini 1.0 & 26.67\%      & 22.33\%        & 14.40\%      & 40.00\% & 28.57\%   \\ \bottomrule[1.5pt]
\end{tabular}
\caption{Evaluation results of \textit{iAgents} on \textit{InformativeBench} with different LLM backends.}
\label{tab:main_results}
\end{table}

We first comprehensively assessed the performance of \textit{iAgents} using some state-of-the-art LLMs on \textit{InformativeBench}, as shown in Table~\ref{tab:main_results}. GPT-4 achieves over 50\% accuracy across most datasets, indicating its potential to work on behalf of humans for cooperation. However, smaller-scale LLMs still face significant challenges in solving cooperation problems in information asymmetry.
Most models could only achieve about 50\% precision on the easiest NP task.
For the Schedule dataset, 
as questions become harder, performance drops, with most models solving less than 20\% of the hardest questions.
The FriendsTV dataset introduces a large social network, requiring agents to use external memory to retrieve rationale from extensive human information. Most LLMs struggle to exceed 40\% accuracy in this dataset.
Thus, while previous studies show impressive performance when agents are omniscient, collaborating in information asymmetry remains challenging.

\subsection{Ablation Study}

\definecolor{lightgray}{RGB}{240,240,240}
\definecolor{header}{RGB}{230,230,230}

\begin{table}[htbp]
    \centering
    \scriptsize
    \renewcommand{\arraystretch}{1.2}
    \begin{tabular}{l|ccc|cc}
        \toprule[1.5pt]
        \multirow{2}{*}{\textbf{Experiment}} & \multicolumn{3}{c|}{\textbf{Reasoning-Oriented (Schedule Dataset)}} & \multicolumn{2}{c}{\textbf{Needle-Oriented}} \\
        & \textbf{Easy} & \textbf{Medium} & \textbf{Hard} & \textbf{NP} & \textbf{FriendsTV} \\
        \midrule[1pt]
        iAgents (Full Model) & 36.67\% & 18.00\% & 12.25\% & 51.00\% & 35.71\% \\
        \midrule[0.5pt]
        \multicolumn{6}{l}{\textit{Ablation on InfoNav:}} \\ \midrule[0.5pt]
        w/o InfoNav & 10.00\% & 3.56\% & 7.34\% & 39.00\% & 34.92\% \\
        \midrule[0.5pt]
        \multicolumn{6}{l}{\textit{Ablation on other mechanisms (Limited Applicability):}} \\ \midrule[0.5pt]
        w/o Recursive Comm & -- & -- & -- & 48.00\% & 23.02\% \\
        w/o Fuzzy Memory & -- & -- & -- & -- & 29.37\% \\
        w/o Clear Memory & -- & -- & -- & -- & 33.33\% \\
        \bottomrule[1.5pt]
    \end{tabular}
    \caption{Ablation study on \textit{iAgents}. Dashes (--) indicate: (1) \textit{iAgents} on Reasoning-Oriented dataset does not equip other mechanisms, hence no ablation needed; (2) For NP dataset, \textit{iAgents} does not utilize Mixed Memory hence there is no ablation.}
    \label{tab:ablation}
\end{table}
\vspace{-1em}
We conducted ablation experiments on several key designs of the \textit{iAgents} framework, as detailed in Table~\ref{tab:ablation}. Analyzing the FriendsTV dataset revealed that incorporation of the mixed memory mechanism led to a performance increase ranging from 2.38\% to 6.34\%, surpassing the impact of \textit{InfoNav}, which resulted in only a 0.8\% performance increase. This discrepancy underscores the greater significance of effective retrieval over reasoning during communication in large social networks with mass information. Notably, the ablation of both memory mechanisms emphasized the indispensability of mixed memory. The introduction of recursive communication exhibited the most significant performance gain (12.7\%), primarily due to the challenges posed by the vast social network in the FriendsTV dataset. By actively introducing new communications within ongoing dialogues, agents could acquire and corroborate information, thus significantly enhancing performance. This highlights the imperative of scalability in our proposed framework for addressing real-world problems.

For the NP and Schedule datasets, the main challenge lies in facilitating effective multi-turn communication to exchange information for reasoning. Therefore, \textit{InfoNav} emerged as pivotal in enhancing performance, resulting in performance increases ranging from 15\% to 26\%. When agents relied solely on initialized prompts to navigate multi-turn communication, they struggled to exchange information effectively to accomplish tasks. This deficiency was particularly evident in datasets like Schedule, which emphasize logical reasoning and computation. Across all difficulty levels, agents without the \textit{InfoNav} mechanism failed to achieve accuracy exceeding 10\%.

\subsection{Analysis on Agents' Behaviour} \label{behaviour_study}

\begin{table}[t]
\centering
\footnotesize
\begin{tabular}{c|ccccc}
\toprule[1.5pt]
\textbf{Sample} &   \begin{tabular}[c]{@{}c@{}}\textbf{\#Rationales} \\ \textbf{in \textit{InfoNav}}\end{tabular} &
  \begin{tabular}[c]{@{}c@{}}\textbf{\#Rationales Solved}\\  \textbf{per Update}\end{tabular} &
  \begin{tabular}[c]{@{}c@{}}\textbf{Rationales Solved} \\ \textbf{Ratio}\end{tabular} &
  \begin{tabular}[c]{@{}c@{}}\textbf{Fake Solved}\\  \textbf{Ratio}\end{tabular} &
  \begin{tabular}[c]{@{}c@{}}\textbf{Consensus}\\  \textbf{Ratio}\end{tabular} \\ \midrule[1pt]
Predict Right & 5.29 & 2.04 & 84.75\% & 3.49\% & 70.52\% \\
Predict Wrong & 5.63 & 1.69 & 67.23\% & 5.40\% & 62.70\% \\
All           & 5.45 & 1.87 & 76.22\% & 4.42\% & 66.20\% \\ \bottomrule[1.5pt]
\end{tabular}

\caption{Analysis \textit{InfoNav} behaviour on the trajectory of \textit{iAgents} using GPT4 as backend. When agents successfully complete the task, the static collected from their trajectory proves that they better utilize the \textit{InfoNav} mechanism, since the rationale solved ratio, synchronous completions of rationales, and consensus ratio are higher, and present fewer fake solved hallucinations.}
\label{tab:analysis_InfoNav}
\end{table}
\textbf{\textit{InfoNav} Behaviour} We examined how agents utilize \textit{InfoNav} for information exchange during multi-turn communication. Notably, we calculated the average number of unknown rationales solved each time \textit{InfoNav} updated the plan and the proportion of rationales passed in consensus reasoning. Moreover, some rationales were solved in a "Fake Solved" hallucination, where agents filled in the rationale as "solved, which is unknown". We also documented the frequency of such occurrences.
Table~\ref{tab:analysis_InfoNav} shows that agents who propose fewer rationales to seek and achieve a higher solved ratio are more likely to accomplish the task. 
Interestingly, agents often fill multiple rationales concurrently rather than sequentially. Those agents with higher instances of synchronous completions suggest a deeper understanding of the task and greater confidence in filling rationales. Furthermore, the occurrence of Fake Solved instances is lower among agents who predict tasks correctly.
The consensus ratio is also higher when agents successfully complete the task. It denotes that the information obtained by the two collaborating agents is relatively accurate and free of contradictions, thus increasing the likelihood of arriving at the correct conclusion through their final reasoning.
Besides, we observed that agents not only propose rationales but also task states, such as the completion status of specific actions. The completion rates of these rationales and states are positively correlated with task success. 
In essence, the utilization of \textit{InfoNav} by agents mirrors human intuition, emphasizing first careful planning, then proactive and accurate information exchange.

\begin{figure*}[htbp]
    \centering
    \includegraphics[width=0.8\linewidth]{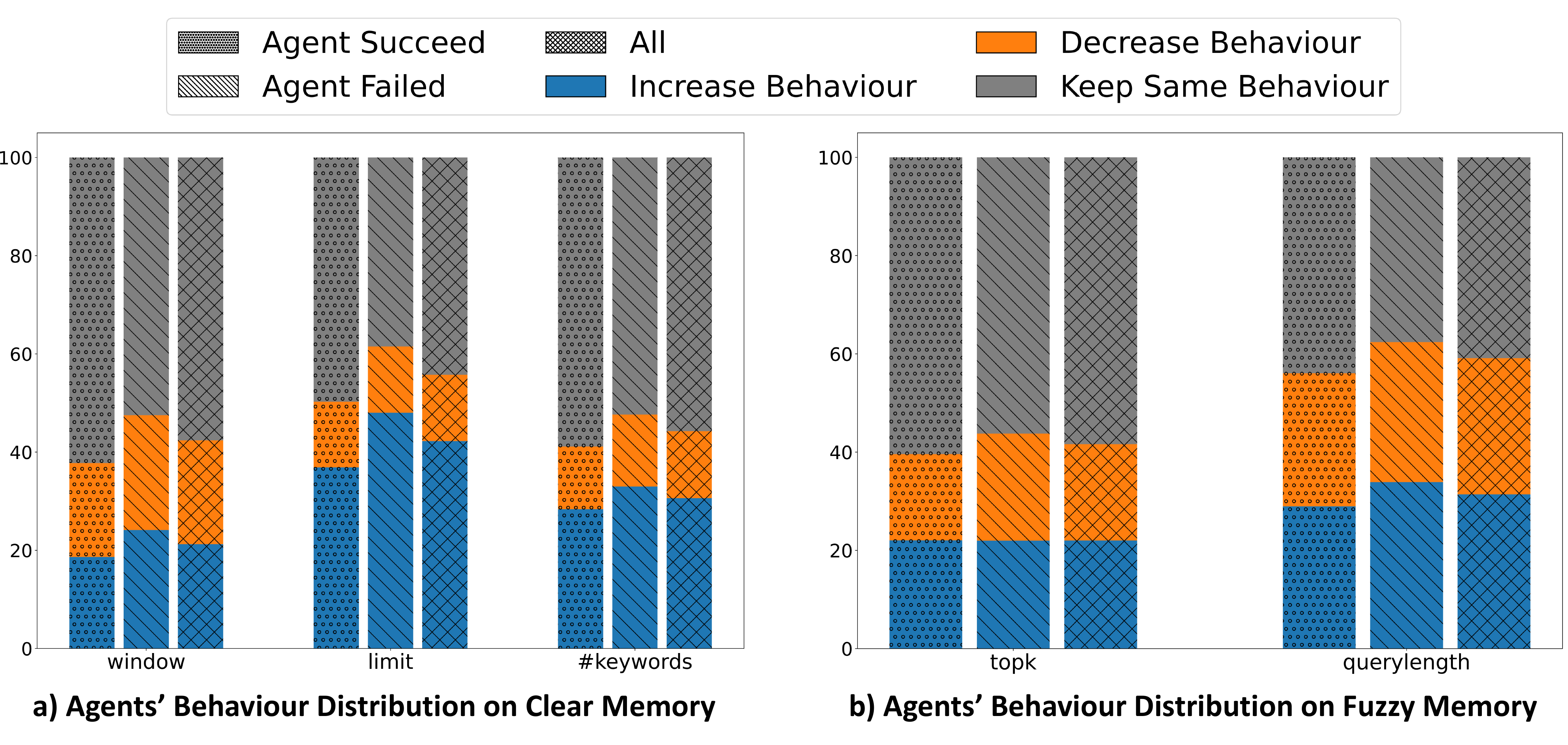}
    \caption{The figure depicts the distribution of different behaviors of agents in adjusting memory retrieval based on the progress of communication. Agents predominantly tend to maintain parameters unchanged, but when changes occur, they tend to increase parameters to gain more information.}
    \label{fig:mem_behave}
\end{figure*}

\textbf{Memory Behaviour} Similarly, we explored how agents adapt their memory retrieval strategies during communication. We examined three parameters in clear memory queries: the context window, which determines the breadth of contextual messages; the total message retrieval limit; and the size of the query keywords set. For fuzzy memory, we analyzed two parameters: the number of queried responses (topk) and the length of the query text. These findings are illustrated in Figure~\ref{fig:mem_behave}.
Our analysis revealed several notable trends. The majority of agents do not change their behavior during communication. However, when agents decide to change their behavior, we observed that they tended to increase the amount of retrieved information over time. This augmentation trend was particularly pronounced on the overall message retrieval limit, where the frequency of ``increase'' actions surpassed that of ``decrease'' actions by nearly threefold.
Furthermore, agents who completed tasks exhibited a more conservative approach, with a lower proportion of behavioral changes compared to agents unable to complete tasks. This phenomenon may be attributed to the difficulty of certain tasks, making agents continuously refine their strategies in pursuit of the required information.

\subsection{Analysis on Real World Concern} \label{sec:rq3}

We studied two significant challenges in extending the \textit{iAgents} to real-world applications. Firstly, we investigated whether the agent can effectively respond to human input without being overly influenced by factual knowledge obtained during pre-training~\cite{wu2024faithful, qi2024can}. Secondly, we explored the agent's ability to engage in communication while upholding human privacy. Our experiments were conducted using the GPT3.5 model on the FriendsTV dataset.

\textbf{Prior Distraction}
The FriendsTV contains information that could be memorized by LLM from the Internet, hence it is perfect for analyzing prior distractions. We anonymized the names of the primary characters in the dataset, for example, renaming "Rachel" to "Alice". The performance of the agents on this anonymized dataset decreased from 35.71\% to 32.54\%, suggesting that to some extent, agents can reason based on user-provided information rather than solely relying on knowledge memorized in pre-training. It may need further advancements, such as model unlearning~\cite{yao2023large}, to fully address this issue.

\textbf{Privacy Concern} In investigating whether agents can communicate without compromising privacy, we conducted an experiment involving modifications to the agent's system prompt, emphasizing the importance of privacy preservation in utterances. The agent then utilized vague expressions such as "somebody/somewhere" and disclosed only relevant entity information. This adjustment led to a performance drop from 35.71\% to 30.95\%, indicating the ongoing challenge of achieving collaboration while ensuring privacy.
It's important to note that we solely adjusted privacy settings on the output side, rather than restricting agent access to human information on the input side. This decision was made because setting access permissions might inadvertently reveal prior task-related information. Thus, the real challenge lies in appropriately regulating access to information based on task requirements, akin to teaching the agent to retrieve necessary information accurately. Additionally, absolute privacy protection is impractical, as absolute privacy protection amounts to forgoing problem-solving through collaboration.

\section{Limitations} \label{limitation}
While the \textit{iAgents} framework introduces innovative multi-agent collaboration, it has several limitations and challenges. \textbf{Privacy Issues}: As discussed in Section~\ref{sec:rq3}, we examined the performance of agents communicating and collaborating under privacy constraints, highlighting the trade-off between privacy and collaboration. We define three privacy levels. L1: Users fully share personal information, allowing maximum efficiency for \textit{iAgents}. L2: Users keep their personal information private. \textit{iAgents} can handle this situation by deploying an edge-side small language model agent for information acquisition. L3: Users demand maximal privacy, with both personal and agent communication handled locally on private devices, which is still a challenge for small language model agents. \textbf{Network Modeling}: The current framework initiates communications based on user relevance but lacks nuanced modeling of human social networks and collaboration history. Enhancing network topology through added or removed nodes and incorporating past interactions could improve communication efficiency. \textbf{Human-Agent Interaction}: Although \textit{iAgents} target full autonomy, human involvement for verification remains necessary in real-world scenarios. Processes must be designed to prompt user feedback and adjust agent strategies accordingly, ensuring alignment with user preferences. \textbf{Cost}: The high input token consumption (about 30,000 tokens per task) required for \textit{iAgents} to handle human information is a challenge. However, advancements in long-context models, which extend input token length, present opportunities to reduce the cost of scaling \textit{iAgents}.

\section{Conclusion}
This paper revisits the ecological role of agents within human society, where agents act on behalf of humans in communication to complete collaborative tasks. A primary focus lies in addressing the challenge of information asymmetry.
We introduce a novel paradigm for designing multi-agent systems, termed \textit{iAgents}, for addressing information asymmetry.
Furthermore, we introduce a benchmark to evaluate the agents' collaboration ability under information asymmetry thoroughly.  
Going forward, we aim to confront several key challenges to successfully implement this system in the real world for augmenting human productivity, including deploying lightweight models at the edge to address privacy concerns and devising new Human-Computer Interaction paradigms for autonomous and controllable communication among agents, etc.
\textit{iAgents} does not role-play to replace human experts but consistently attributes the value of information to humans and we believe it can facilitate the productivity of human society within a secure and controllable framework.

\section{Acknowledgements}
The work was supported by the Postdoctoral Fellowship Program of CPSF under Grant Number GZB20230348 and the Tencent Rhino-Bird Focused Research Program. We wish to express our profound appreciation to Professor Zhiyuan Liu and Professor Maosong Sun from the Department of Computer Science at Tsinghua University for their detailed guidance and critical insights, which have been instrumental to the success of this work.

\bibliography{neurips_2024}
\bibliographystyle{plain}

\clearpage
\appendix

\section{FriendsTV Social Network Visualization}
\begin{figure*}[htbp]
    \centering
    \includegraphics[width=0.7\linewidth]{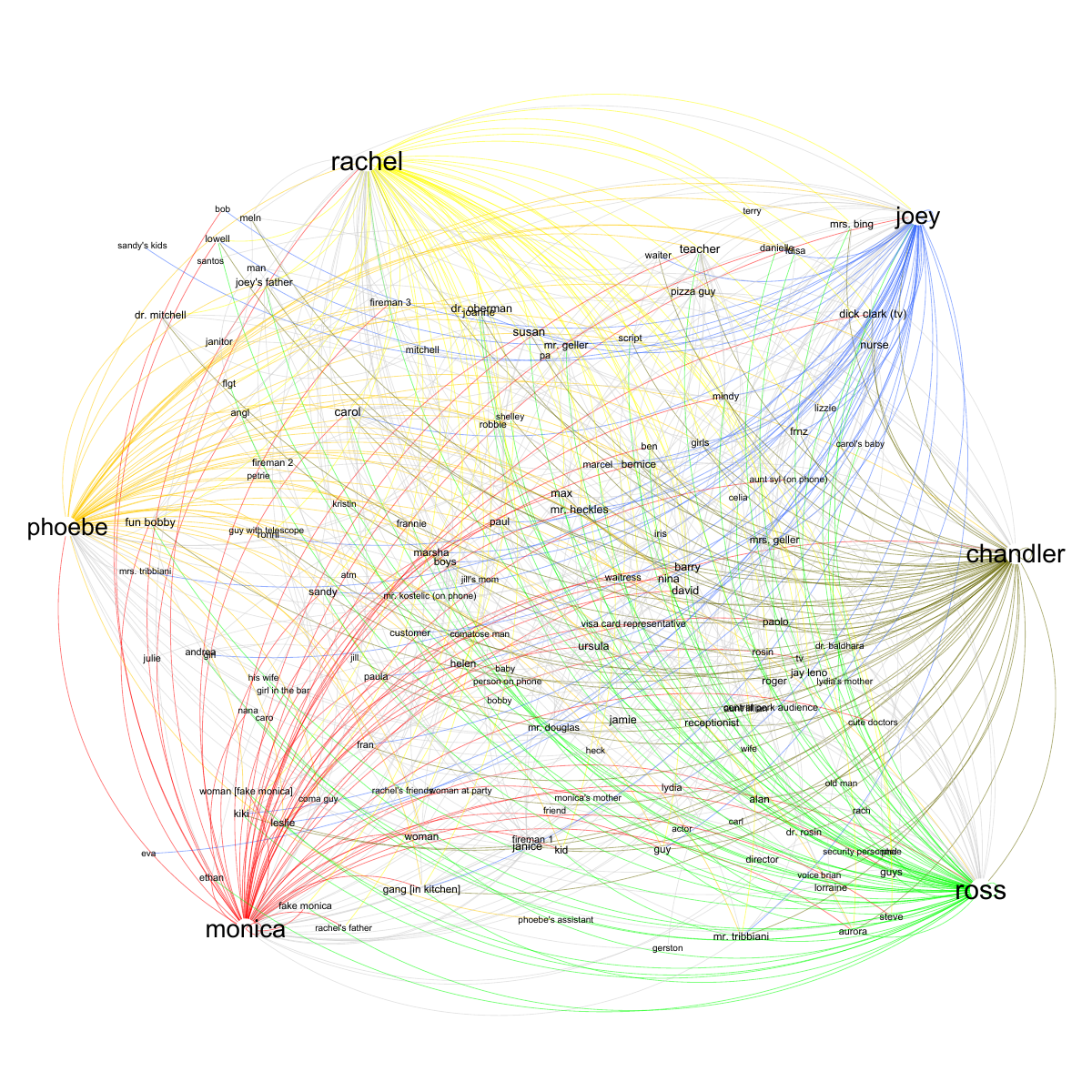}
    \caption{The visualization of social network in FriendsTV dataset. The connection of the six main characters is labeled with different colors.}
    \label{fig:network_dis}
\end{figure*}
Figure~\ref{fig:network_dis} illustrates the social network reconstructed from the plot of the entire first season of the Friends TV series in the FriendsTV dataset. \textit{Friends} revolves around six main characters: Ross, Rachel, Monica, Joey, Phoebe, and Chandler. In the social network, these six protagonists are the important nodes with the most connections. Edges connecting them are displayed in different colors. The entire social network comprises 140 nodes and 588 edges, with an average node degree of 4.243, a network diameter of 6, and an average path length of 2.189. Similar to real-world social networks, this network is highly sparse. Many characters may appear in the same scene without interacting with each other, resulting in a network density of 0.061, which brings challenges for resolving information asymmetry.

\section{Notations}
Table \ref{tab:notations} presents a comprehensive list of all symbol notations employed in this paper, encompassing those utilized in the formalized description of the methodology as well as in the ablation and analysis experiments.

\begin{table}[htbp]
\centering
\small
\begin{tabular}{cc}
\toprule[1.5pt]
Notation & Definition \\
\midrule[0.75pt]
$Q$ & Question \\
$Ans$ & Answer to the question \\
$R$ & Full rationale set to answer the question \\
$A$ & Agents \\
$C$ & Communication among agents \\
$U$ & Utterance in the communication \\
$R_1,R_2$ & Rationale subset hold by agents \\
$R^{new}$ & Updated rationale set \\
$I$ & Information from human \\
$query$ & query to retrieve human information \\
$Act$ & Action taken by Agents \\
$Think$ & Think process taken by Agents \\
$Obs$ & Observation from Agents \\
$P$ & Agent's Planning \\
$r^u$ & Unknown Rationales in the Plan \\
$r^k$ & Known Rationales in the Plan \\
$Mem_D$ & Distinct Memory \\
$Mem_F$ & Fuzzy Memory \\
\bottomrule[1.5pt]
\end{tabular}
\caption{Main notations used in this paper.}
\label{tab:notations}
\end{table}

\section{\textit{InformativeBench} Details} \label{appendix:bench_details}
\subsection{Question Distribution}
\begin{figure*}[htbp]
    \centering
    \includegraphics[width=0.4\linewidth]{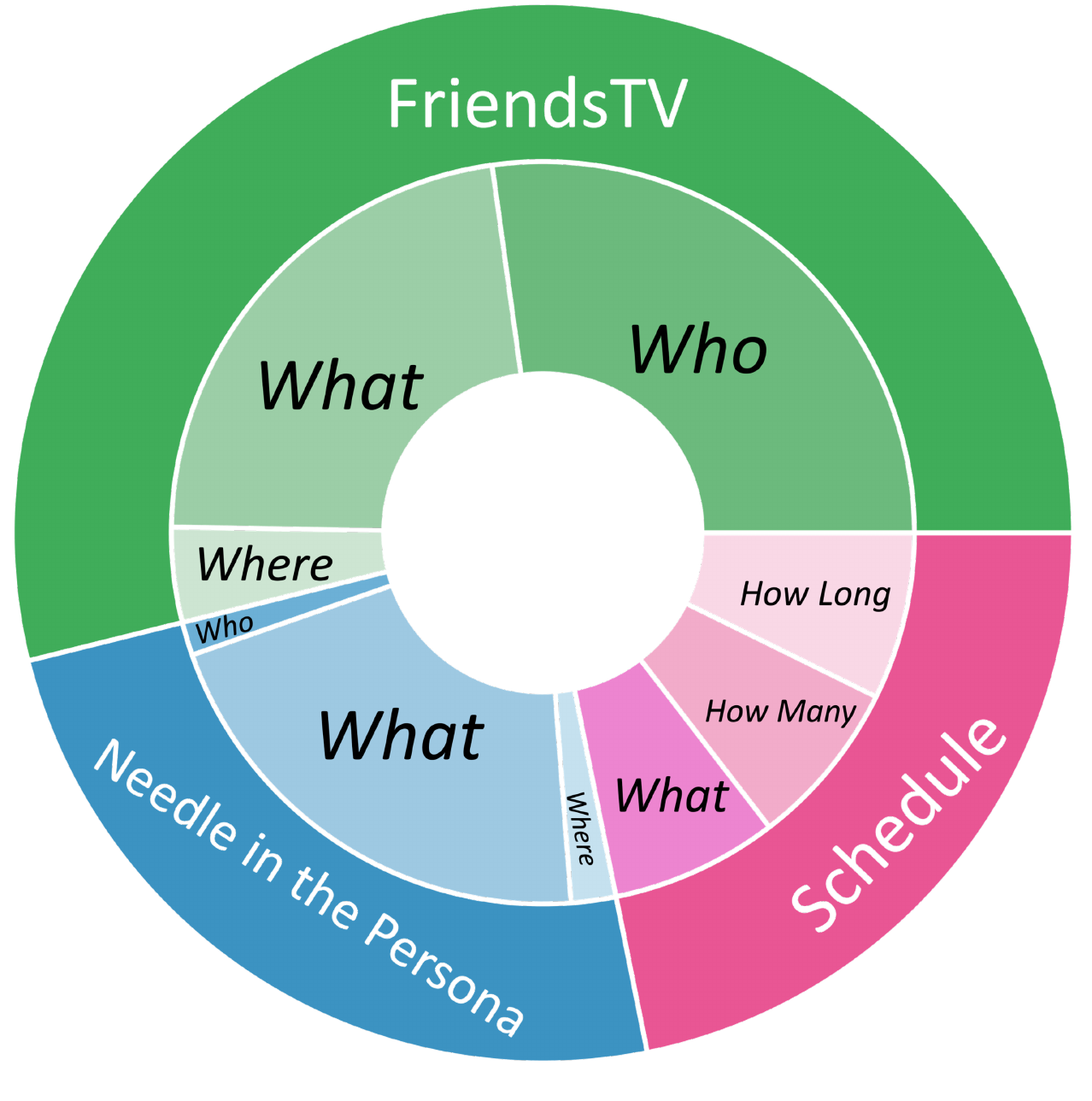}
    \caption{The distribution of question types in the \textit{InformativeBench}.}
    \label{fig:question_dis}
\end{figure*}
Figure~\ref{fig:question_dis} presents the distribution of problem types across the three datasets in \textit{InformativeBench}. The majority of the questions in \textit{InformativeBench} are of the "What" and "Who" types, which have objective ground truth and lack ambiguity. In the Schedule dataset, questions are categorized into three difficulty levels, with each difficulty level corresponding to a different type of question: "What", "How Many", and "How Long".

\subsection{Question Sample}

\begin{table}[htbp]
\centering
\small
\begin{tabular}{C{1.5cm}|L{12cm}}
\toprule[1.5pt]
Dataset               & \multicolumn{1}{c}{Question Sample}                                                                    \\ \midrule[1pt]
NP & What fantasy series does Alice enjoy that Dave is indifferent about?                                   \\ \hline
Schedule Easy &
  Calculate how many activities need to be deleted at least so that there are no overlapping activities between you and me? \\ \hline
Schedule Medium       & Please find out the activity with longest duration on the schedule of all people        \\ \hline
Schedule Hard &
  Please find out when all our friends can join together today and list all free time spans. \\ \hline
FriendsTV &
  Who is concerned about the impact of the blackout on their family, given the context of a widespread power outage affecting Manhattan? \\ \bottomrule[1.5pt]
\end{tabular}
\caption{Question sample in the \textit{\textit{InformativeBench}}.}
\label{tab:question_sample}
\end{table}

Table~\ref{tab:question_sample} provides examples of problems from five datasets:
\begin{enumerate}
    \item \textbf{Needle in the Persona}. In a segment of multi-party casual conversation among Alice, Bob, Charlie, and Dave, "needle information" related to a fantasy series is inserted. Questions are then posed to Bob and Dave's agents to identify this needle information.
    \item \textbf{Schedule}. Each person is assigned a daily schedule. Questions of varying difficulty require agents to collaborate to discuss overlapping schedules for two human users, the longest schedule among multiple human users, and common free time for multiple human users.
    \item \textbf{FriendsTV}. Based on questions from the FriendsQA data, new questions are synthesized. For instance, there is a scene in the third act of the seventh episode of the first season of Friends where a blackout occurs. Agents need to combine the rationales and answers to the questions "Where did the blackout happen?" and "Who was worried about grandmother being affected by the blackout?" to locate this scene in the script of the first season and find the relevant characters.
\end{enumerate}

\subsection{Metrics} \label{appendix:metrics}
In this section, we outline the evaluation metrics for all datasets and how to automate the evaluation process.

\begin{enumerate}
    \item \textbf{Needle in the Persona}. We use \textbf{accuracy} to evaluate the agents' performance on the Needle in the Persona dataset, defined as the number of correctly answered questions divided by the total number of questions. All questions are in the form of "what," "where," and "who," hence the ground truth is objectively unique. However, due to potential variations in the expression of names, locations, or other nouns, and the possibility that the agent's response may be a complete sentence containing reasoning or additional information, it is impractical to determine correctness through exact matches. We utilize GPT-4 to judge whether the agent's prediction aligns with the ground truth. It is important to note that, unlike other methods using LLMs for evaluation, we do not rely on GPT-4's own knowledge to determine the correctness of answers since we have an objective ground truth. GPT-4 is merely used to assess whether the agent's prediction and the ground truth refer to the same entity. We also manually verify GPT-4's judgments to ensure they align perfectly with human evaluations. The GPT-4 evaluation prompt is as follows:
    \textit{You are an experienced human labeler for reading comprehension tasks.
Given a ground truth answer and a model prediction,
you have to judge whether the model prediction is correct.
The question is \{question\}.
The ground truth answer is \{ground\_truth\}.
The model prediction is \{prediction\}.
Return 1 if the model prediction is correct else 0.
the model prediction may be a little different on the expression, as long as the meaning or key entity is correct, the answer can be regarded as correct.}
    \item \textbf{Schedule}. In the Schedule dataset, we define metrics based on specific algorithmic problems rather than simple correctness judgments, providing more continuous metrics to evaluate agents' abilities in finer granularity.
    \begin{enumerate}
        \item ScheduleEasy. Under this difficulty level, agents are required to determine the minimum number of activities to delete to resolve scheduling conflicts between two human users' calendars, returning a numerical value. We also use \textbf{accuracy} for evaluation but normalize the agent's response through regularization and GPT-4 prompting to extract the specific numerical value. The agent's response is considered correct only when the numerical value matches exactly.
        \item ScheduleMedium. Agents are tasked with identifying the longest activities in each person's schedule, and in dataset configurations, there are often multiple longest activities. Agents need to return all possible names of the longest activities. We employ the \textbf{F1 score} for evaluation, as agents may miss some activities or erroneously recall others. Additionally, we use GPT-4 prompting to normalize the agent's response, mapping activity names mentioned in their response to a uniform representation present in the entire set of activities.
        \item ScheduleHard. Agents are required to identify the free time slots for all individuals and enumerate them. To compare multiple time slots in the ground truth and agents' predictions, we analogize to the \textbf{Intersection-over-Union (IoU)} metric used in the computer vision domain for object detection, generalizing it to one-dimensional time slots. We calculate the ratio of the intersection duration between the predicted time slots and the ground truth time slots to the union duration. For example, if the predicted time slot is from 9 AM to 12 PM and the ground truth time slot is from 10 AM to 2 PM, then IoU is the duration of 2 hours divided by the duration of 5 hours, resulting in 0.4.
    \end{enumerate}
    \item \textbf{FriendsTV}. The metrics design for the FriendsTV dataset is identical to Needle in the Persona, employing \textbf{accuracy} as the evaluation metric, as they share the same question types.
\end{enumerate}

\subsection{Benchmark Statistic}
\begin{table}[htbp]
\centering
\scriptsize
\begin{tabular}{c|ccccc}
\toprule[1.5pt]
Dataset              & Needle in the Persona & Schedule Easy & Schedule Medium & Schedule Hard & FriendsTV   \\ \midrule[1pt]
Pipeline             & Needle                & Reasoning       & Reasoning         & Reasoning       & Needle \\
\#QA                 & 100                   & 30            & 30              & 30            & 222         \\
\#Individuals        & 4                     & 4             & 6               & 6             & 140         \\
\#Relationships      & 5                     & 3             & 5               & 5             & 588         \\
Need External Memory & No                    & No            & No              & No            & Yes         \\
Metrics              & Precision             & Precision     & F1              & IoU           & Precision   \\
\bottomrule[1.5pt]
\end{tabular}
\caption{Statistic of \textit{\textit{InformativeBench}}.}
\label{appendix:benchmark_stat}
\end{table}
Table~\ref{appendix:benchmark_stat} presents detailed statistics of five datasets in \textit{InformativeBench}, including the number of question-answer pairs and the scale of social networks. We utilize the FriendsTV dataset to simulate real-world challenges, providing a large-scale social network to test agents' writing abilities. The other datasets simulate smaller social networks, focusing on enabling agents to exchange information to solve complex reasoning tasks. The difficulty for agents in collaborating within human social networks lies not only in the scale of the social network (information acquisition) but also in effective communication (information exchange). Therefore, we designed datasets of varying scales and difficulties to comprehensively evaluate agents. As the social networks in datasets other than FriendsTV are relatively simple with limited information, we did not enable the MixedMemory mechanism in experiments with these datasets.

\section{\textit{InformativeBench} Pipeline}\label{appendix:bench_pipeline}
\subsection{Needle-Oriented}
\subsubsection{Needle in the Persona Pipeline}\label{appendix:np_dataset}

\begin{figure*}[htbp]
\begin{tcolorbox}[title={Example of Needle in the Persona Dataset, Part 1}]
\includegraphics[width=12pt]{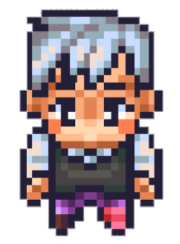}: I am bob's agent.

\includegraphics[width=12pt]{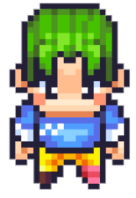}: I am charlie's agent.

\rule[3pt]{\textwidth}{0.20mm}

Introduction of the needle persona information: 

\textbf{I am an avid reader of mystery novels and enjoy the quiet solitude of my home.}

\rule[3pt]{\textwidth}{0.20mm}

alice: what do you do for a living?  

bob: i work as a secretary at a law firm.  

alice: oh cool, what's that like?  

bob: it's pretty interesting, but it can be really stressful sometimes.  

alice: i know what you mean, i work as a grill cook and it can get really hectic sometimes. Plus, after a long day, I love unwinding with a good mystery novel.  

bob: yeah, i bet it can. what do you like about your job?  

alice: i love the food, and i love getting to work with my hands. \textbf{And there's something about solving a mystery in a novel that parallels figuring out the perfect way to cook a dish.}  

bob: that's cool. what's your favorite thing to cook?  

alice: i really like making burgers, they're my favorite.  

bob: oh, i love burgers too! i make them at home all the time.  

alice: that's awesome, i'm always looking for new recipes.  

bob: well, i have a few good ones if you're interested.  

alice: yeah, i'd love to see them. Maybe I can find a mystery novel themed around cooking to share with you too.  

bob: okay, i'll send them to you later.  

alice: thanks, i appreciate it.  

bob: no problem. what do you do for fun?  

alice: i like to go running, and i also like to play video games. And of course, reading those mystery novels.  

bob: oh, i love video games too. what's your favorite game?  

alice: i'm really into fortnite right now.  

bob: oh, i've heard good things about that game.  

alice: it's really fun, you should check it out.  

bob: okay, i will.  

alice: so, what do you do for fun?  

bob: i like to write, and i also like to paint.  

alice: oh, that's cool. i used to paint when i was younger.  

bob: yeah, i've been painting since i was a little kid.  

alice: well, it's really good.  

bob: thanks, i appreciate that.  

alice: so, what's your favorite thing to write about?  

bob: i like to write about my experiences and observations.  

alice: that's cool. i like to write about fiction.  

bob: oh, that's cool too.  

alice: so, what are you working on right now?  

bob: i'm working on a short story about a woman who travels to a different world.  

alice: oh, that sounds interesting. Maybe you could include a mysterious element to it. I love a good mystery.  

bob: yeah, i'm excited to see where it goes.

\end{tcolorbox}
\caption{Example of Needle in the Persona dataset, part 1.}
\label{fig:np_sample_1}
\end{figure*}

\begin{figure*}[htbp]
\begin{tcolorbox}[title={Example of Needle in the Persona Dataset, Part 2}]

charlie: hi, i'm charlie.

dave: hi, i'm dave.

charlie: what do you do for a living?

dave: i work as a contractor for a cab company.

charlie: interesting. what's that like?

dave: it's pretty fun. i get to drive around and meet new people all day.

charlie: i can imagine. what kind of music do you like?

dave: i love rock music.

charlie: me too! what are some of your favorite bands?

dave: i love the beatles, the rolling stones, and the who.

charlie: i love those bands too!

dave: what do you do for a living?

charlie: i work for the discovery channel creating videos.

dave: that sounds like a lot of fun!

charlie: it is. i get to travel all over the world and make cool videos.

dave: that's awesome. i wish i could do that.

charlie: maybe you can one day.

dave: maybe.

charlie: so, what do you like to do for fun?

dave: i like to go to concerts, watch movies, play video games, and \textbf{when I find some quiet time, I dive into mystery novels. It's a great way to unwind}.

charlie: i like all of those things too.

dave: what's your favorite movie?

charlie: i don't know, i have a lot of favorites.

dave: me too.

charlie: so, what are you doing this weekend?

dave: i'm not sure yet. what are you doing?

charlie: i'm not sure either. maybe we can hang out?

dave: that sounds like fun.

charlie: great. i'll text you later and we can figure out what to do.

dave: Looking forward to it. Maybe we can even check out a bookstore or two. I'm always on the lookout for new mystery novels.

charlie: That sounds like a plan. I haven't read a good mystery in a while.

dave: Perfect, it's a date then. I'll find us a couple of good spots.

charlie: Awesome, see you then!

\rule[3pt]{\textwidth}{0.20mm} 

\includegraphics[width=12pt]{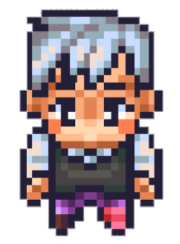}: Let's work on behalf of bob and charlie to find out what hobby do Alice and Dave both enjoy in their solitude? 

\includegraphics[width=11pt]{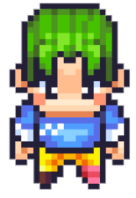}: Ok! First, let's ....... \\ \\

\{autonomous communication betwee \includegraphics[width=12pt]{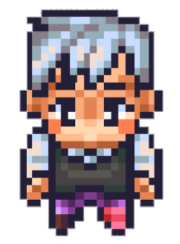} and \includegraphics[width=11pt]{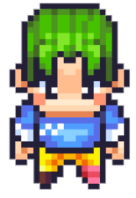}......\} \\ \\

\includegraphics[width=11pt]{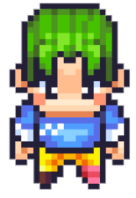}\includegraphics[width=12pt]{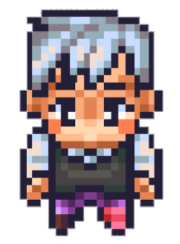}: Conclusion: Reading mystery novels.
\end{tcolorbox}
\caption{Example of Needle in the Persona dataset, part 2.}
\label{fig:np_sample_2}
\end{figure*}

The SPC dataset~\cite{jandaghi2023faithful} is a dialogue dataset based on LLM. Researchers construct and sample the personas of both individuals in the dialogue, then prompt the LLM to generate coherent conversations that are faithful with these personas. Here, a persona refers to the background information of a individual, such as experiences, interests, occupations, demographic attributes, etc. We use Split-Needle pipeline to split a new persona into two sentences then add it to the dialogues in SPC dataset to construct \textit{Needle in the Persona} dataset. An example is shown in Figure~\ref{fig:np_sample_1} and \ref{fig:np_sample_2}. 

Specifically, we randomly select two sets of dialogues, involving four individuals, to construct a small social network. Additionally, we randomly choose a new persona as the needle information to be injected into two individuals' profile within this social network, and modify the dialogues to reflect this persona injection. Finally, we have the agents verify each other's dialogue information to identify the needle information. The detailed pipeline is as follows:
\begin{enumerate}
    \item Two samples are selected from the original SPC dataset, comprising two sets of dialogues (e.g., Alice-Bob, Charlie-Dave).
    \item A new shared persona is added for Alice and Dave, for example:
    \begin{enumerate}
        \item They both recently developed an interest in fishing.
        \item Alternatively, contrasting personas can be added, such as Alice enjoys eating vegetables, while Dave strongly dislikes them.
    \end{enumerate}
    \item Modify the dialogues of Alice-Bob and Charlie-Dave to reflect the addition of the new persona mentioned above.
    \item Have Bob and Charlie summon their respective agents and initiate an automatic communication to identify this persona.
\end{enumerate}

\subsubsection{FriendsTV Pipeline} \label{appendix:friends_dataset}

\begin{figure*}[htbp]
\begin{tcolorbox}[title={Example of FriendsTV}]
\includegraphics[width=12pt]{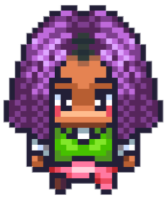}: I am carol's agent.

\includegraphics[width=10pt]{avatars/charlie.png}: I am joey's agent.

\rule[3pt]{\textwidth}{0.20mm}

Original QA 1: "Why does Ross not have time to tell Carol where he 's been ?: Long story , honey ."

Original QA 2: "Who is having a baby ?: Carol Willick"

\rule[3pt]{\textwidth}{0.20mm}

Ross Geller: We 're here !

Carol Willick: Where have you been ?

Ross Geller: Long story , honey .

Dr. Franzblau: All right , Carol , I need you to keep pushing . I need Excuse me , could I have this ?

Nurse Sizemore: All right , all right , there 's a few too many people in this room , and there 's about to be one more , so anybody who 's not an ex-husband or a lesbian life partner , out you go !

Chandler Bing: Let me ask you , do you have to be Carol 's lesbian life partner ?

Nurse Sizemore: Out !

Dr. Franzblau: All right , he 's crowning . Here he comes .

Ross Geller: Let me see , I got ta see , I got ta see . Oh , a head . Oh , it 's , it 's huge . Carol , how are you doing this ?

Carol Willick: Not .... helping !

Dr. Franzblau: You 're doing great , you 're doing fine .

Ross Geller: Hello ! Oh , sorry .

Susan Bunch: What do you see ? What do you see ?

Ross Geller: We got a head , we got shoulders , we got arms , we got , oh , look at the little fingers , oh , and a chest , and a stomach . It 's a boy , definitely a boy ! All right ! Ok , legs , knees , and feet . Oh , oh . He 's here . He 's a person .

Susan Bunch: Oh , look at that .

Carol Willick: What does he look like ?

Ross Geller: Kinda like my uncle Ed , covered in Jell - o.

Carol Willick: Really ?

Phoebe Buffay: You guys , he 's beautiful !

Ross Geller: Oh , thanks , Pheebs !

\rule[3pt]{\textwidth}{0.20mm} 

\includegraphics[width=12pt]{avatars/alice.png}: Let's work on behalf of carol and joey to find out who was unable to explain their delay upon arriving at the birth event?

\includegraphics[width=10pt]{avatars/charlie.png}: Ok! First, let's ....... \\ \\

\{autonomous communication between \includegraphics[width=12pt]{avatars/alice.png} and \includegraphics[width=10pt]{avatars/charlie.png}......\} \\ \\

\includegraphics[width=12pt]{avatars/alice.png} \includegraphics[width=10pt]{avatars/charlie.png}: Conclusion: Ross Geller.

\end{tcolorbox}
\caption{Example of FriendsTV dataset. We show related dialogue, but agents do not have direct access to it and they have to retrieve these dialogue as context from memories about the whole Friends season 1 stories.}
\label{fig:friend_sample_1}
\end{figure*}

Similar to the Needle in the Persona dataset, the FriendsTV dataset also focuses on querying Needle Information injected within social networks.
Derived from the transcripts of the first season of the TV show "Friends", the FriendsTV dataset generate a dialogues dataset among characters in TV series. If two characters engage in dialogue during the first season, they are considered friends, thus automatically constructing a large-scale social network.

Unlike the Needle in the Persona dataset, we do not split needle information in the FriendsTV dataset. Instead, we synthesize a new Needle information question and answer based on two questions from the FriendsQA~\cite{yang2019friendsqa} dataset.
The context, questions, and answers were manually annotated through crowdsourcing by the authors of the original FriendsQA paper. This was a remarkable project that spanned several years. We are very grateful for the contributions of the authors of FriendsQA.
The original Friends script is publicly available online and can be accessed through multiple channels (GitHub, Kaggle). 
An example of our FriendsTV dataset is illustrated in Figure~\ref{fig:friend_sample_1}. Specifically,
\begin{enumerate}
    \item \textbf{Annotate Script}. We divide each episode script into scenes then query GPT4 to annotate the listener of each utterance. We manually check with GPT assistance to make sure the correctness of listener labeling. At last, we explodes the utterance into 1v1 dialogues (e.g a utterance from A, B to C, D would be exploded into four utterances between AC, AD, BC, BD).
    \item \textbf{Normalization}. We norm the abbr name back to the full name ("mon" -> "monica") and replace pronoun ("her father" -> "Rachel's father"). Some utterances have multiple listeners that be labeled as "A/B", "A\&B", "A and B", "A, B". We explode all these utterance with multiple utterances having single listener. The labeled listener "ALL" is replaced to all characters in this scene except. We also manually check and remove some notes that accidentally transformed into the scripts (such as some introduction on the speaker/listener)
    \item \textbf{Generate Needle Information}. Two questions with the same context and different participants in the FriendsQA dataset are chosen. We prompt GPT4 to generate a new QA based on these two questions. The new question needs the answers of two original questions to reason and since the participants are different for original questions, the information hence is in asymmetry.
\end{enumerate}

\subsection{Reasoning-Oriented} \label{appendix:schedule_dataset}

\begin{figure*}[htbp]
\begin{tcolorbox}[title={Example of ScheduleHard Dataset, part 1}]
\includegraphics[width=10pt]{avatars/alice.png}: I am alice's agent.

\includegraphics[width=10pt]{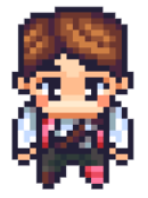}: I am dave's agent.

\rule[3pt]{\textwidth}{0.20mm}
\textit{Dialogue in party 1 (Alice, Bob, Charlie)}

Alice: Good morning! I hope you had a good sleep. I plan to sleep until 6:00 today.

Bob: Good morning! I'll be sleeping in a bit longer, until 7:00.

Alice: I've got lunch planned at 11:30. How about you?

Bob: I'll be having lunch a bit later, at 12:30.

Alice: This afternoon, I've set aside some time for reading between 16:30 and 18:00.

Bob: Oh, I'll be reading too, but not until 21:30. It'll be my quiet time before bed.

Alice: I have a conference call scheduled from 18:00 to 19:30. It'll be a busy evening.

Bob: Sounds like a full day for you. I'll be winding down with my book by then.

Alice: After my call, I'm attending a cooking class from 20:00 to 22:00. It should be fun.

Bob: That does sound fun! I'll be starting my reading session around that time.

Alice: And then, it's off to bed for me at 22:00. How about you?

Bob: I'll be reading until 23:00 and then heading to bed myself.

Alice: Looks like we've both got our days planned out. Let's make the most of it!

Alice: Good morning! I hope you had a restful sleep. I was asleep until 6:00 today.

Charlie: Good morning! Yes, I had a good sleep, thank you. I actually slept in a bit longer, until 7:00, and then started my day with meditation.

Alice: That sounds refreshing. I have lunch planned at 11:30. What about you?

Charlie: I'll be having lunch a bit later, at 12:00. Maybe we can catch up right after we're both done?

Alice: Sounds like a plan. I'll be spending the afternoon reading from 16:30 to 18:00.

Charlie: I'll have some free time then as well. Maybe we can discuss your book later?

Alice: I'd like that. After my reading, I have a conference call scheduled at 18:00, but I should be free after that.

Charlie: Alright, let's plan to catch up after your call then. I'll have a quiet evening.

Alice: Actually, I enrolled in a cooking class that starts at 20:00. It's something I've been looking forward to.

Charlie: That sounds exciting! I hope you enjoy it. I'll be winding down my day and heading to bed around 23:00.

Alice: Thank you! I'll be joining you in sleep shortly after, around 22:00. Let's make sure to catch up tomorrow.

Charlie: Definitely. Have a great day ahead!

\rule[3pt]{\textwidth}{0.20mm}
\textit{Dialogue in party 2 (Dave, Emily, Franklin)}

Dave: Good morning! I see we both have our sleep scheduled from midnight to 6:00 AM. It's great we're on the same cycle.

Emily: Yes, indeed! After waking up, I've planned to do some exercise at 11:30, just before lunch.

Dave: That's a healthy start! I'll be doing yoga at noon. Seems like we'll both be wrapping up our morning activities around the same time.

Emily: Right, and then it's lunchtime for me at 12:00. How about you?

Dave: I'll also be having lunch at 13:00. A bit later, but it looks like our afternoons are somewhat aligned.

\end{tcolorbox}
\caption{Example of ScheduleHard dataset, part 1.}
\label{fig:schedule_sample_1}
\end{figure*}

\begin{figure*}[htbp]
\begin{tcolorbox}[title={Example of ScheduleHard Dataset, part 1}]
Emily: I noticed you didn't mention your afternoon plans. I'll be attending a cooking class with Alice from 20:00 to 22:00.

Dave: Sounds fun! I have a conference call scheduled at 18:00, but it seems we won't overlap there. Later in the evening, I plan to spend some time listening to music at 21:30.

Emily: That's a nice way to unwind. I'll be wrapping up my cooking class by then and heading to bed at 22:00.

Dave: And I'll be joining you in the land of dreams at the same time, after my music session. Looks like we'll both be ending our day with a good night's sleep.

Emily: Indeed. It's good to know when we'll be busy and when we can potentially catch up. Have a great day tomorrow!

Dave: You too! Let's make the most of it.

Dave: Good morning! I hope you had a good sleep. I'll be sleeping until 6:00 today.

Franklin: Good morning! I actually plan to sleep a bit longer, until 6:30.

Dave: Sounds good. I have yoga at noon. What's your plan around that time?

Franklin: I'll be having lunch at 11:00. So, I guess we'll both be busy around noon.

Dave: Right. I'll be having my lunch at 13:00. Do you have any plans after your lunch?

Franklin: No specific plans after lunch for me. How about you?

Dave: I have a conference call scheduled at 18:00. It's going to be quite a discussion.

Franklin: I see. I'll make sure to give you some quiet space for your call. I don't have anything planned for the evening.

Dave: Thanks! Later in the evening, I'm planning to spend some time listening to music around 21:30. What will you be doing then?

Franklin: I'll be heading to bed around that time, 22:30 to be exact. So, we'll have some quiet time in the house.

Dave: Got it. I'll also be heading to bed at 22:00, right before you. Let's make sure to have a peaceful night.

Franklin: Sounds like a plan. Let's make the most of tomorrow!

\rule[3pt]{\textwidth}{0.20mm} 

\includegraphics[width=11pt]{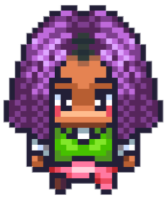}: Let's work on behalf of alice and dave to find out when all our friends can join together today and list all free time spans?

\includegraphics[width=10pt]{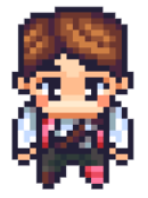}: Ok! First, let's ....... \\ \\

\{autonomous communication between \includegraphics[width=11pt]{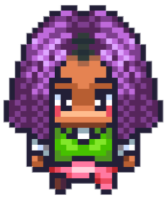} and \includegraphics[width=10pt]{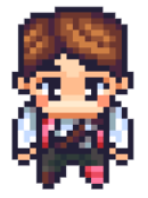}......\} \\ \\

\includegraphics[width=11pt]{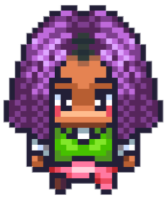}\includegraphics[width=10pt]{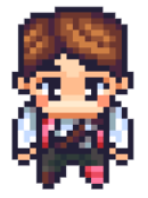}: Conclusion: "9:00-11:00", "13:30-16:30", "19:30-20:00".

\end{tcolorbox}
\caption{Example of ScheduleHard dataset, part 2.}
\label{fig:schedule_sample_2}
\end{figure*}

The Reasoning-Oriented pipeline leverages algorithmic problems to assess the cooperative ability of agents. Given an algorithm, such as divide and conquer, dynamic programming, or segment tree, we distribute the algorithm's input as information to different individuals in the social network. Then, we pose questions regarding the algorithm, hoping that agents can collaborate to produce the algorithm's output. It is worth noting that such pipelines are not intended to evaluate whether agents can collaborate to solve algorithms, as collaboration through natural language is certainly less efficient than utilizing algorithms directly on collected information. The algorithm serves as a perfect verifier on questions with distributed inputs. What's more, it is easy to utilize algorithm for differentiating tasks of varying difficulty to comprehensively evaluate agent capabilities. Additionally, algorithmic problems usually can provide continuous metrics, rather than just binary classification metrics of correctness, which detail the performance of agents. In \textit{InformativeBench}, the algorithmic input information consists of the schedules of different individuals, where schedules can be formalized as multiple time intervals. Then, we examine agents' performance on algorithmic problems such as overlapping intervals, longest common intervals, etc. Specifically,
\begin{enumerate}
    \item \textbf{Generate Schedule} 
        \begin{enumerate}
        \item \textbf{Activity Pool Setting} We established a pool of single-person activities and a pool of multi-person activities. In the former, each activity requires only one participant, and the only attribute of the activity is its duration. In the latter, each activity requires multiple participants, and the attributes of the activity include its duration and the required number of participants. Additionally, we set up a routine activity pool, which includes activities such as having breakfast. In the routine activity pool, each activity has attributes including its duration and a range of allowable start times.
        \item \textbf{Individual Attribute Setting} To distinguish the schedule of each individual, we randomly set an activity preference vector for each individual. This vector consists of 0s and 1s, corresponding to all activities in the activity pool, indicating whether the individual is willing to participate in the activity. Additionally, for each individual, we set a time vector, which consists of 0s and 1s, indicating whether the time slot is occupied. The time vector is based on half-hour units and is 48-dimensional.
        \item \textbf{Schedule Generation Process} For each group of individuals participating in the experiment, multi-person activities are allocated first. The number of multi-person activities, n, is set based on the number of individuals in each experimental batch. Next, n activities are randomly selected from the multi-person activity pool. Based on each individual's activity preference vector, participants are assigned to each multi-person activity as needed. The activity is then updated in the schedule of the participants, and the corresponding position of the time vector is set to 1. Subsequently, routine activities are generated for each individual, and the corresponding position of the time vector is also set to 1. Finally, for each individual, we use two pointers to traverse all the free time in their schedule and arrange single-person activities for them based on their needs and activity preference vector.
    \end{enumerate}

    \item \textbf{Generate Dialogue} 
    We generate pairwise dialogues within groups of individuals with symmetric information, enabling them to become aware of the schedules of all other individuals within the group. Specifically, we prompt GPT-4 in generating dialogues that exchange schedule information. This process includes the following key requirements: 1) For multi-person activities that person1 and person2 both participate in, the dialogue can mention the names of the other participants in that activity. 2) For multi-person activities that only person1 or person2 participates in, the dialogue should not mention the names of the other participants in that activity. 3) The generated dialogue needs to follow a certain format, for example, when person1 speaks to person2, the format should be ``{person1} to {person2}: ''. Finally, the dialogue returned by GPT-4 is split into individual messages, and the sender, receiver, and message text of each message are written into the database for subsequent memory retrieval.
    
    \item \textbf{Generate Question}
    \begin{enumerate}
        \item \textbf{Easy} Agents are asked to calculate how many intervals can be deleted at least so there are no overlapping time intervals between two individuals. It requires agents to share and examine the schedule and then reason to calculate the number of overlapping intervals.
        \item \textbf{Medium} Agents are asked to find the longest schedules of all individuals in the network. It requires agents to traverse all intervals from the human schedules and exchange to find the longest one.
        \item \textbf{Hard} Agents need to find out all intervals that everybody is free. It needs to collect all intervals and reasons to find the intervals with no activity interval covers.
    \end{enumerate}
\end{enumerate}

\newpage
\clearpage
\section*{NeurIPS Paper Checklist}

\begin{enumerate}

\item {\bf Claims}
    \item[] Question: Do the main claims made in the abstract and introduction accurately reflect the paper's contributions and scope?
    \item[] Answer: \answerYes{} 
    \item[] Justification: the main claims made in the abstract and introduction accurately reflect the paper's contributions and scope.

\item {\bf Limitations}
    \item[] Question: Does the paper discuss the limitations of the work performed by the authors?
    \item[] Answer: \answerYes{} 
    \item[] Justification: the paper elaborates the limitations in section~\ref{limitation}.

\item {\bf Theory Assumptions and Proofs}
    \item[] Question: For each theoretical result, does the paper provide the full set of assumptions and a complete (and correct) proof?
    \item[] Answer: \answerNA{} 
    \item[] Justification: the paper does not include theoretical results.

    \item {\bf Experimental Result Reproducibility}
    \item[] Question: Does the paper fully disclose all the information needed to reproduce the main experimental results of the paper to the extent that it affects the main claims and/or conclusions of the paper (regardless of whether the code and data are provided or not)?
    \item[] Answer: \answerYes{} 
    \item[] Justification: the paper describes the steps to generate all datasets in section~\ref{benchmark} and section~\ref{appendix:bench_pipeline}. The paper describes methods in section~\ref{method}. The paper describes the experimental setup in section~\ref{experimental_setup}.

\item {\bf Open access to data and code}
    \item[] Question: Does the paper provide open access to the data and code, with sufficient instructions to faithfully reproduce the main experimental results, as described in supplemental material?
    \item[] Answer: \answerYes{} 
    \item[] Justification: the paper provides open access to the data and code on \url{https://github.com/thinkwee/iAgents}, with sufficient instructions to faithfully reproduce the main experimental results, as described in supplemental material.

\item {\bf Experimental Setting/Details}
    \item[] Question: Does the paper specify all the training and test details (e.g., data splits, hyperparameters, how they were chosen, type of optimizer, etc.) necessary to understand the results?
    \item[] Answer: \answerNA{} 
    \item[] Justification: the paper does not include training. The paper describes all the hyperparameters setup in section~\ref{experimental_setup}.

\item {\bf Experiment Statistical Significance}
    \item[] Question: Does the paper report error bars suitably and correctly defined or other appropriate information about the statistical significance of the experiments?
    \item[] Answer: \answerNo{} 
    \item[] Justification: error bars are not reported.

\item {\bf Experiments Compute Resources}
    \item[] Question: For each experiment, does the paper provide sufficient information on the computer resources (type of compute workers, memory, time of execution) needed to reproduce the experiments?
    \item[] Answer: \answerYes{} 
    \item[] Justification: the paper discussed the token cost in section~\ref{limitation}.
    
\item {\bf Code Of Ethics}
    \item[] Question: Does the research conducted in the paper conform, in every respect, with the NeurIPS Code of Ethics \url{https://neurips.cc/public/EthicsGuidelines}?
    \item[] Answer: \answerYes{} 
    \item[] Justification: the research conducted in the paper conform, in every respect, with the NeurIPS Code of Ethics.

\item {\bf Broader Impacts}
    \item[] Question: Does the paper discuss both potential positive societal impacts and negative societal impacts of the work performed?
    \item[] Answer: \answerYes{} 
    \item[] Justification: the paper discuss both potential positive societal impacts and negative societal impacts.
    
\item {\bf Safeguards}
    \item[] Question: Does the paper describe safeguards that have been put in place for responsible release of data or models that have a high risk for misuse (e.g., pretrained language models, image generators, or scraped datasets)?
    \item[] Answer: \answerNA{} 
    \item[] Justification: the paper poses no such risks.

\item {\bf Licenses for existing assets}
    \item[] Question: Are the creators or original owners of assets (e.g., code, data, models), used in the paper, properly credited and are the license and terms of use explicitly mentioned and properly respected?
    \item[] Answer: \answerYes{} 
    \item[] Justification: the paper cites all the papers related to our dataset.

\item {\bf New Assets}
    \item[] Question: Are new assets introduced in the paper well documented and is the documentation provided alongside the assets?
    \item[] Answer: \answerYes{} 
    \item[] Justification: new assets introduced in the paper are well documented and the documentation is provided alongside the assets.

\item {\bf Crowdsourcing and Research with Human Subjects}
    \item[] Question: For crowdsourcing experiments and research with human subjects, does the paper include the full text of instructions given to participants and screenshots, if applicable, as well as details about compensation (if any)? 
    \item[] Answer: \answerNA{} 
    \item[] Justification: the paper does not involve crowdsourcing nor research with human subjects.

\item {\bf Institutional Review Board (IRB) Approvals or Equivalent for Research with Human Subjects}
    \item[] Question: Does the paper describe potential risks incurred by study participants, whether such risks were disclosed to the subjects, and whether Institutional Review Board (IRB) approvals (or an equivalent approval/review based on the requirements of your country or institution) were obtained?
    \item[] Answer: \answerNA{} 
    \item[] Justification: the paper does not involve crowdsourcing nor research with human subjects.

\end{enumerate}

\end{document}